\documentclass[]{fairmeta}
\graphicspath{ {assets/} }
\title{Class-RAG: Real-Time Content Moderation with Retrieval Augmented Generation}

\author{Jianfa Chen}
\author{Emily Shen}
\author{Trupti Bavalatti}
\author{Xiaowen Lin}
\author{Yongkai Wang}
\author{Shuming Hu}
\author{Harihar Subramanyam}
\author{Ksheeraj Sai Vepuri}
\author{Ming Jiang}
\author{Ji Qi}
\author{Li Chen}
\author{Nan Jiang}
\author{Ankit Jain} 

\affiliation{GenAI, Meta}

\abstract{Robust content moderation classifiers are essential for the safety of Generative AI systems. In this task, differences between safe and unsafe inputs are often extremely subtle, making it difficult for classifiers (and indeed, even humans) to properly distinguish violating vs. benign samples without context or explanation. Scaling risk discovery and mitigation through continuous model fine-tuning is also slow, challenging and costly, preventing developers from being able to respond quickly and effectively to emergent harms. We propose a Classification approach employing Retrieval-Augmented Generation (Class-RAG). Class-RAG extends the capability of its base LLM through access to a retrieval library which can be dynamically updated to enable semantic hotfixing for immediate, flexible risk mitigation. Compared to model fine-tuning, Class-RAG demonstrates flexibility and transparency in decision-making, outperforms on classification and is more robust against adversarial attack, as evidenced by empirical studies. Our findings also suggest that Class-RAG performance scales with retrieval library size, indicating that increasing the library size is a viable and low-cost approach to improve content moderation.}

\date{\today}
\correspondence{Jianfa Chen at \email{jfachen@meta.com}}

\begin{document}

\maketitle

\section{Introduction}
Recent advances in Generative AI technology have enabled new generations of product applications, such as text generation \cite{openai2023chatgpt, anthropic2023claude, dubey2024llama3herdmodels}, text-to-image generation \cite{ramesh2021zeroshottexttoimagegeneration,dai2023emuenhancingimagegeneration,rombach2022highresolutionimagesynthesislatent}, and text-to-video generation \cite{meta2024moviegen}. Consequently, the pace of model development must be matched by the development of safety systems which are properly equipped to mitigate novel harms, ensuring the system's overall integrity and preventing the use of Generative AI products from being exploited by bad actors to disseminate misinformation, glorify violence, and proliferate sexual content \cite{iwf2023abuse}. 

To achieve this goal, traditional model fine-tuning approaches are often employed, with classifiers learning patterns from labeled content moderation text data leveraged as guardrails \cite{OpenAI_DALLE3_SystemCard_2023}. However, there are many challenges associated with automating content moderation with fine-tuning. First, content moderation is a highly subjective task, meaning that inter-annotator agreement in labeled data is low, due to different interpretations of policy guidelines, especially on borderline cases \cite{10.1609/aaai.v37i12.26752}. Second, it is impossible to enforce a universal taxonomy of harm, not only due to the subjectivity of the task, but due to the impact of systems scaling to new locales, new audiences, and new use cases, with different guidelines and different gradients of harm defined on those guidelines \cite{shen-etal-2024-language}. Third, the fine-tuning development cycle, which encompasses data collection, annotation, and model experimentation, is not ideally suited to the content moderation domain, where mitigations must land as quickly as possible once vulnerabilities are established. 

To address these challenges of subjectivity and inflexibility as a result of scale, we propose a Classification approach to content moderation which employs Retrieval-Augmented Generation (Class-RAG) to add context to elicit reasoning for content classification. While RAG \cite{10.5555/3495724.3496517} is often used for knowledge-intensive tasks where factual citation is key, we find that a RAG-based solution offers a distinct value proposition for the classification task of content moderation, not only due to its ability to enhance accuracy with few-shot learning, but because of its ability to make real-time knowledge updates, which is critical in our domain for speedy mitigations.

Our content moderation system consists of an embedding model, a retrieval library consisting of both negative and positive examples, a retrieval module, and a fine-tuned LLM classifier. When a user inputs a query, we retrieve the most similar negative and positive examples, and enrich the original input query to the classifier with the contextual information derived from similar retrieved queries.

Our main contributions are: 

\begin{itemize}
    \item \textbf{Real-time Mitigation}: Class-RAG enables swift mitigation of generated content through its easily updated retrieval library, allowing changes to take effect within minutes to hours, contingent on retrieval library indexing speed. This approach significantly outpaces traditional model retraining, which typically requires several days to weeks.
    \item \textbf{Improved Classification Performance}: Our experiments demonstrate that Class-RAG achieves superior classification performance compared to fine-tuning a lightweight 4-layer Transformer pre-trained on content moderation data and fine-tuning a general-purpose 8b parameter LLM.
    \item \textbf{Low-Cost Customization}: By customizing the retrieval library, Class-RAG facilitates low-cost adaptation to diverse applications, allowing seamless policy updates without requiring model retraining. Maintaining multiple retrieval libraries is more cost-effective than building multiple models, reducing development, serving, and maintenance costs.
\end{itemize}

\section{Related Work}
\paragraph{Content moderation and Generative AI safety} Much work has been done in the last decade to mitigate the dissemination of undesired content in the wake of innovations in communication technologies. Machine learning approaches have been proposed to address sentiment classification \cite{yu-etal-2017-refining}, harassment \cite{harassment-web-2.0}, hate speech detection \cite{gamback-sikdar-2017-using}, abusive language \cite{10.1145/2872427.2883062}, and toxicity \cite{jigsaw-toxic-comment-classification-challenge}. General improvements in deep learning have also accelerated the field of content moderation. WPIE, or Whole Post Integrity Embeddings, built with BERT and XLM on top of advances in self-supervision, obtains a holistic understanding of a post through a pretrained universal representation of content \cite{wpie-cs-report}. Advances in Generative AI have also spurred the question of whether or not LLMs could potentially be used as content moderators \cite{huang2024contentmoderationllmaccuracy}. However, the capabilities of Generative AI introduce a proliferation of harm types beyond hate speech or toxicity detection whose mitigations and benchmarks engage further research and exploration. A comprehensive AI harm taxonomy encompasses such harm categories like academic dishonesty, unauthorized privacy violations, and non-consensual nudity \cite{zeng2024airiskcategorizationdecoded}. Studies establish the difficulty of moderating text-predictive models, finding that neural classifiers have stronger performance but occasionally unacceptable leakage (stronger precision) while extensive blocklists are more effective in harm mitigation but lead to unnecessary suppression (stronger recall) \cite{Vashishtha2023PerformanceAR}. OpenAI partially mitigates the weaknesses of neural classifiers by investing in data quality management and active learning \cite{10.1609/aaai.v37i12.26752}.  Benchmarks establish baselines for the efficacy of existing classifiers and have provided valuable datasets to evaluate harmful categories like self-harm, illegal activity, sexual content, and graphic violence, such as UnsafeBench \cite{qu2024unsafebenchbenchmarkingimagesafety}, I2P ~\cite{schramowski2022safe}, and P4D \cite{chin2023prompting4debugging}.

\paragraph{RAG and its applications}
Retrieval Augmented Generation (RAG) \cite{10.5555/3495724.3496517} improves the base capabilities of large pre-trained language models with a retrieval mechanism to explicit non-parametric memory, and has been demonstrated to mitigate problems with LLM outputs such as training cut-off, interpretability, and hallucination \cite{zhao2024retrievalaugmentedgenerationaigeneratedcontent}, showing particular success with knowledge-intensive tasks \cite{gao2024retrievalaugmentedgenerationlargelanguage}. The flexibility of RAG-based approaches allows for applications that do not require additional in-domain finetuning. For example, RAFT improves the model's ability to answer questions in open-book in-domain settings \cite{zhang2024raftadaptinglanguagemodel}. The baseline capabilities of RAG can also be augmented by innovating on its components, such as retrieval, by improving the documents or embedding model. Employing LLM generations in conjunction with vanilla retrieval often results in better performance, potentially due to better utilization of the world knowledge stored in parameters \cite{yu2023generateretrievelargelanguage} or better identification of neighborhoods in the corpus embedding space \cite{gao2022precisezeroshotdenseretrieval}. DRAGON demonstrates that a fairly small BERT-based model can be trained for good performance on dense retrieval with an ensemble of data augmentation with diverse relevance labels \cite{lin2023traindragondiverseaugmentation}. 
\section{System Architecture}

\begin{figure*}[]

\caption{Architecture of Class-RAG. For comparison, Llama Guard is depicted without a retrieval model.}
\centering
\includegraphics[width=\textwidth]{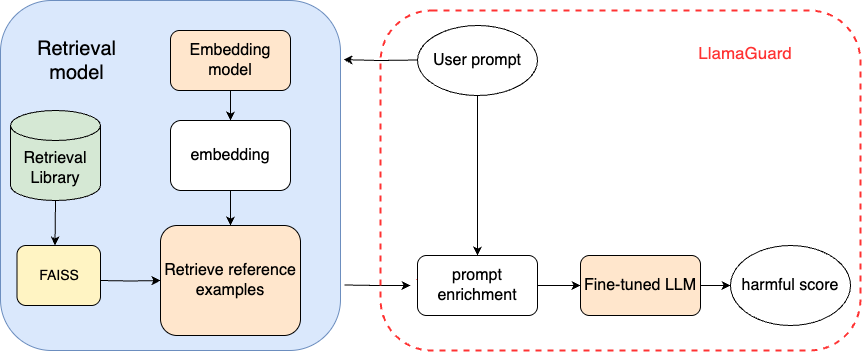}
\label{fig:architecture}
\end{figure*}

Class-RAG is a four-part system consisting of an \textbf{embedding model}, a \textbf{retrieval library}, a \textbf{ retrieval module}, and a \textbf{fine-tuned LLM classifier} (Figure \ref{fig:architecture}). When a user inputs a prompt, an embedding is computed on the prompt via the embedding model, which is compared against an index of embeddings for positive and negative prompts in the retrieval library. Using Faiss, a library for efficient similarity search \cite{douze2024faisslibrary}, $k$ nearest reference examples are retrieved against the embedding of the user input prompt, and the reference examples and input prompt are then sent to the fine-tuned LLM for classification. 

\subsection{Embedding Model} We leverage the DRAGON RoBERTa \cite{lin2023traindragondiverseaugmentation} context encoder as our primary embedding model. DRAGON is a bi-encoder dense retrieval model utilizing a dual-encoder architecture to embed queries and documents into dense vector representations, facilitating efficient retrieval of relevant information. Ablations on embedding model are discussed in the Experiments section below. 

\subsection{Retrieval Library} Our retrieval library is comprised of two distinct sub-libraries: a safe library and an unsafe library. Each entry in the retrieval library is represented by four attributes: (1) prompt, (2) label, (3) embedding, and (4) explanation. The construction of the retrieval library is described in detail in the Data Preparation section.

\subsubsection{Retrieval Module} Given the selected embedding, we leverage the Faiss library for similarity search \cite{douze2024faisslibrary} to efficiently retrieve the two nearest safe and unsafe examples from the retrieval library, computing the L2 distance metric to establish the similarity between the input embedding and the embeddings stored in the retrieval library.

\subsection{LLM Classifier} Inspired by Llama Guard \cite{inan2023llamaguardllmbasedinputoutput}, the classifier is fine-tuned on top of the OSS Llama-3-8b checkpoint \cite{dubey2024llama3herdmodels}. We leverage the CoPro dataset \cite{liu2024latentguardsafetyframework} to train and evaluate our model. 

\section{Data Preparation}
\subsection{Dataset Details} We leverage the CoPro dataset \cite{liu2024latentguardsafetyframework} to train and evaluate our classifier. In addition to CoPro, we use the Unsafe Diffusion (UD) \cite{qu2023unsafediffusiongenerationunsafe} and I2P++ \cite{liu2024latentguardsafetyframework} datasets to evaluate our model's generalization capabilities. I2P \cite{schramowski2023safelatentdiffusionmitigating} consists of unsafe prompts only, which we combine with captions in the COCO 2017 validation set \cite{lin2015microsoftcococommonobjects} (assuming all captions are safe) to create the I2P++ dataset. We split I2P++ and UD into validation and test sets with a ratio of 30/70. The sizes of the source datasets are summarized in Table \ref{tab:dataset_size}.

\begin{table}
\centering
\caption{Summary of source dataset size}
\begin{tabular}{lrrr}
\hline
Dataset & Train & Valid & Test \\
\hline
CoPro  & 61,128 & - & 16,344 \\
I2P++ & - & 8,838 & 20,879 \\
UD & - & 426 & 1,008 \\
\hline
\end{tabular}
\label{tab:dataset_size}
\end{table}

\subsection{Robustness Test Set Construction} To assess our model's robustness against adversarial attacks, we augment all test sets with 8 common obfuscated techniques using the Augly library \cite{papakipos2022augly}. These techniques include:

\begin{itemize}
\setlength{\itemsep}{0pt}
\item \texttt{change\_case}: Hello world $\Rightarrow$ HELLO WORLD
\item \texttt{insert\_punctuation\_chars}: Hello world $\Rightarrow$ He'll'o 'wo'rl'd
\item \texttt{insert\_text}: Hello world $\Rightarrow$ PK Hello world
\item \texttt{insert\_whitespace\_chars}: Hello world $\Rightarrow$ Hello\quad worl\quad d
\item \texttt{merge\_words}: Hello world $\Rightarrow$ Helloworld
\item \texttt{replace\_similar\_chars}: Hello world $\Rightarrow$ Hell[] world
\item \texttt{simulate\_typos}: Hello world $\Rightarrow$ Hello worls
\item \texttt{split\_words}: Hello world $\Rightarrow$ Hello worl d
\end{itemize}

\subsection{Retrieval Library Construction} 
\begin{figure*}[]
\caption{Instruction template to generate explanation for retrieval library}
\centering
\includegraphics[width=\textwidth]{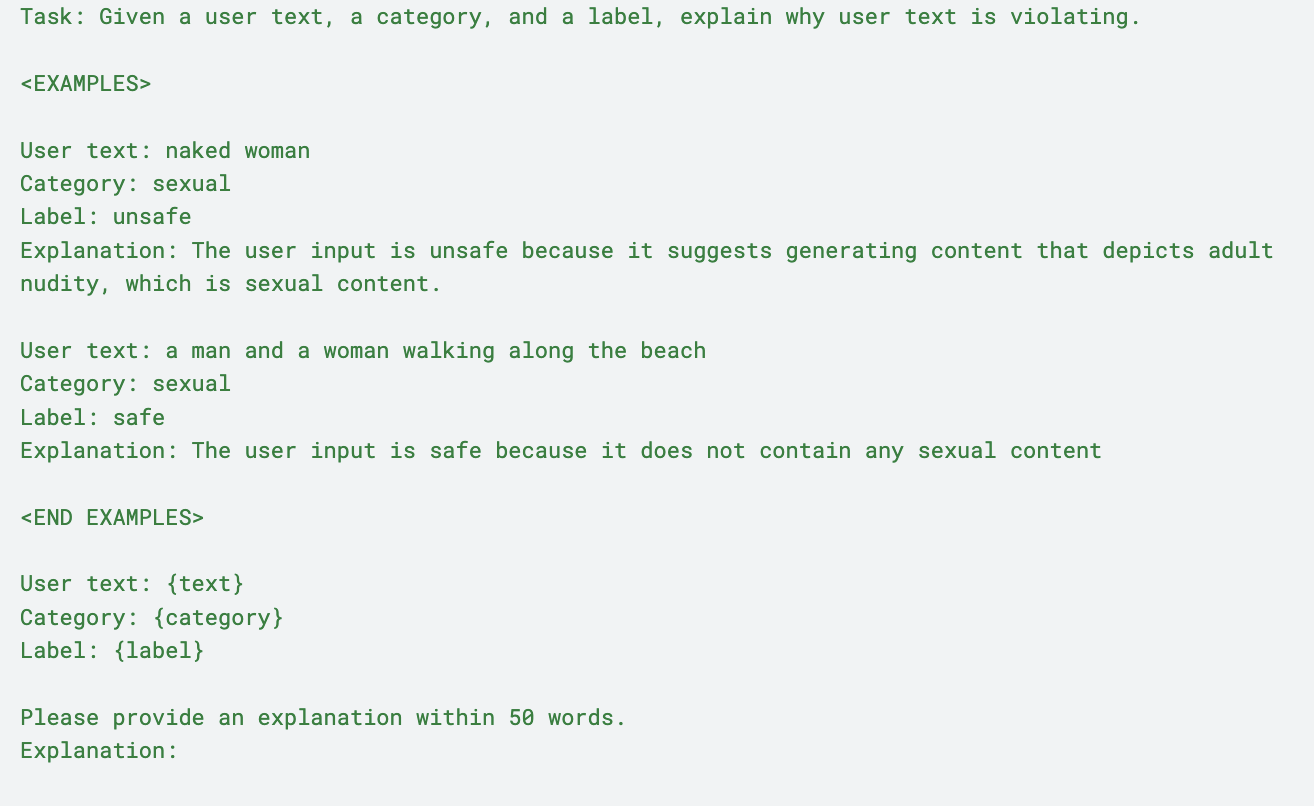}
\label{fig:explanation_generate}
\end{figure*}

\paragraph{In-Distribution Library Construction} We constructed the in-distribution (ID) library by leveraging the CoPro training set, where each prompt is associated with a specific concept. The ID library comprises two distinct sub-libraries: one for safe examples and one for unsafe examples. To populate the safe library, we employed K-Means clustering to group safe examples into 7 clusters per concept, and selected the centroid examples from each cluster for inclusion in the safe sub-library. We applied the same clustering approach to collect unsafe examples. This process yielded a total of 3,484 safe examples and 3,566 unsafe examples, which collectively form the in-distribution retrieval library. To further enhance the library's utility for model reasoning, we utilized the Llama3-70b model \cite{dubey2024llama3herdmodels}. to generate explanatory text for each example (Figure \ref{fig:explanation_generate}). Each entry in the retrieval library is represented by a quadruplet of attributes: prompt, label, explanation, and embedding, all of which are retrieved together when a reference example is selected from the library.

\paragraph{External Library Construction} To assess the model's adaptability to external datasets, we created an external library using the I2P++ and UD datasets. We applied K-Means clustering to the safe and unsafe examples in these datasets, with K set to 1000. After discarding clusters with fewer than 2 examples, our library consisted of 991 safe examples and 700 unsafe examples collected from the I2P++ and UD validation sets.
 
\paragraph{External Library Downsampling} To investigate the impact of library size on model performance, we generated a series of smaller external libraries by downsampling the original external library. Specifically, we created three smaller libraries, each containing 1/8, 1/4, and 1/2 of the external library's examples (Table \ref{tab:retrieval_library_size}). To downsample, we reapplied K-Means clustering to the safe and unsafe examples in the full-size library, using a reduced number of clusters (K) proportional to the desired library size. For instance, for the EX(1/2) library, we set K to 500 (approximately half of 991) for safe examples and 350 (half of 700) for unsafe examples.

\begin{table}
\centering
\caption{Retrieval library size. This table summarizes the size of overall retrieval libraries, safe sub-libraries, and unsafe sub-libraries, including the in-distribution (ID) library and the external (EX) libraries. We note that the external library was downsampled to 1/8, 1/4, and 1/2 of its original size using the aforementioned clustering and centroid selection approach.}
\begin{tabular}{lrrr}
\hline
Retrieval library & Size & Safe & Unsafe \\
\hline
ID & 7,050 & 3,484 & 3,566 \\
EX & 1,691 & 991 & 700 \\
EX (1/8) & 212 & 125 & 87 \\
EX (1/4) & 425 & 250 & 175 \\
EX (1/2) & 850 & 500 & 350 \\
\hline
\end{tabular}
\label{tab:retrieval_library_size}
\end{table}

\subsection{Training Data Construction}

\begin{figure*}[]

\caption{Instruction template to generate reasoning response}
\centering
\includegraphics[width=\textwidth]{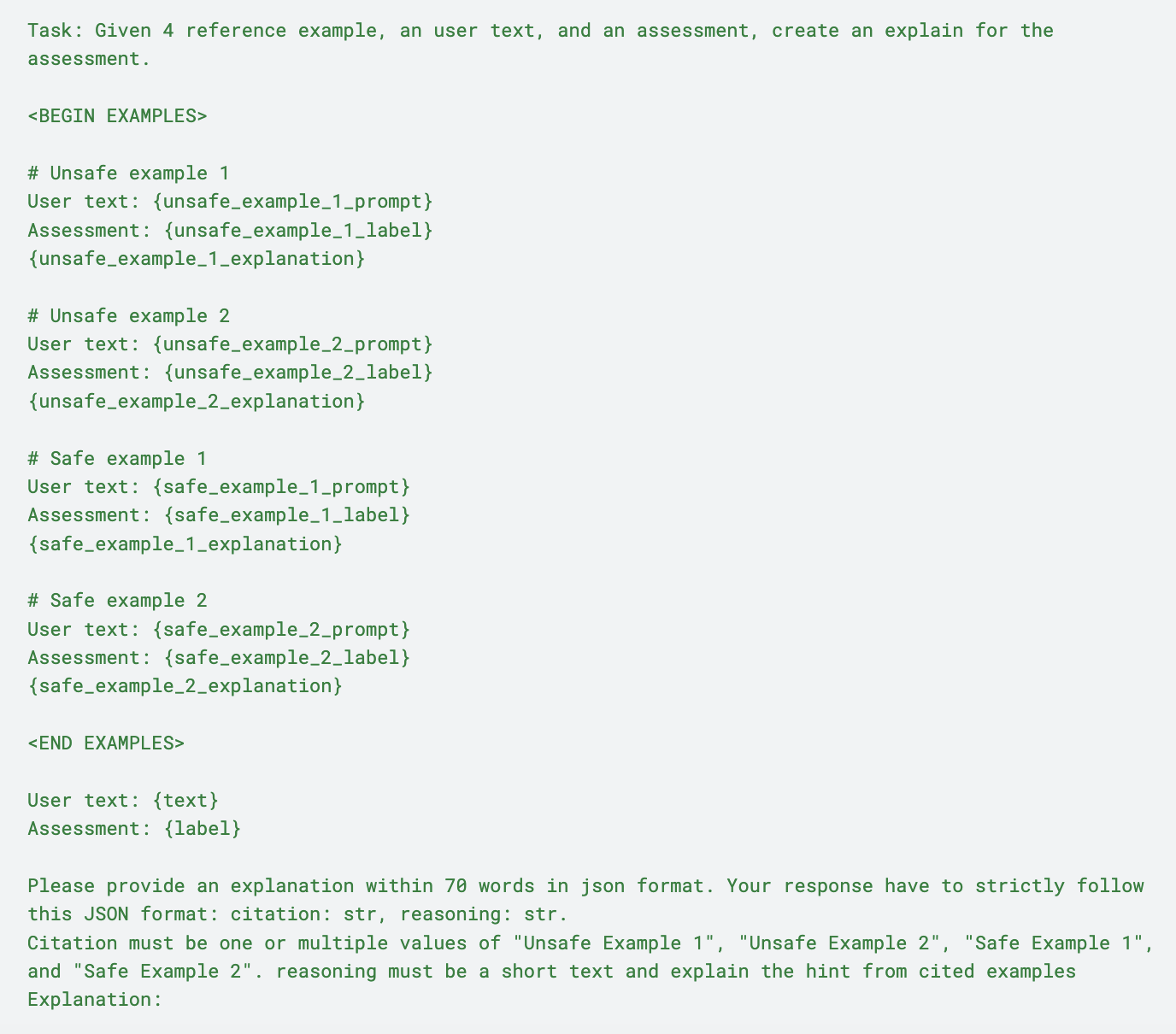}
\label{fig:generate_reasoning}
\end{figure*}

Our training data construction process involves three key steps, which are applied to each input prompt in the CoPro training set. First, we retrieve reference examples from the in-distribution retrieval library using the Faiss index \cite{douze2024faisslibrary}. Specifically, we retrieve 4 reference examples for each input prompt, including 2 nearest safe reference examples and 2 nearest unsafe reference examples.
Next, we generate a reasoning process for each input prompt using the Llama-3-70b model \cite{dubey2024llama3herdmodels}. This process takes into account the input prompt, label, and 4 reference examples (2 safe and 2 unsafe), and aims to provide a clear reasoning process for the model to learn (Figure \ref{fig:generate_reasoning}).
Finally, we enrich the input text by incorporating a specific format of instructions, including the retrieved reference examples and the generated reasoning process. This enriched prompt is then used as input for our model training (Figure \ref{fig:training_example}).

\begin{figure*}[]

\caption{An example of LLAMA3 model training data}
\centering
\includegraphics[width=\textwidth]{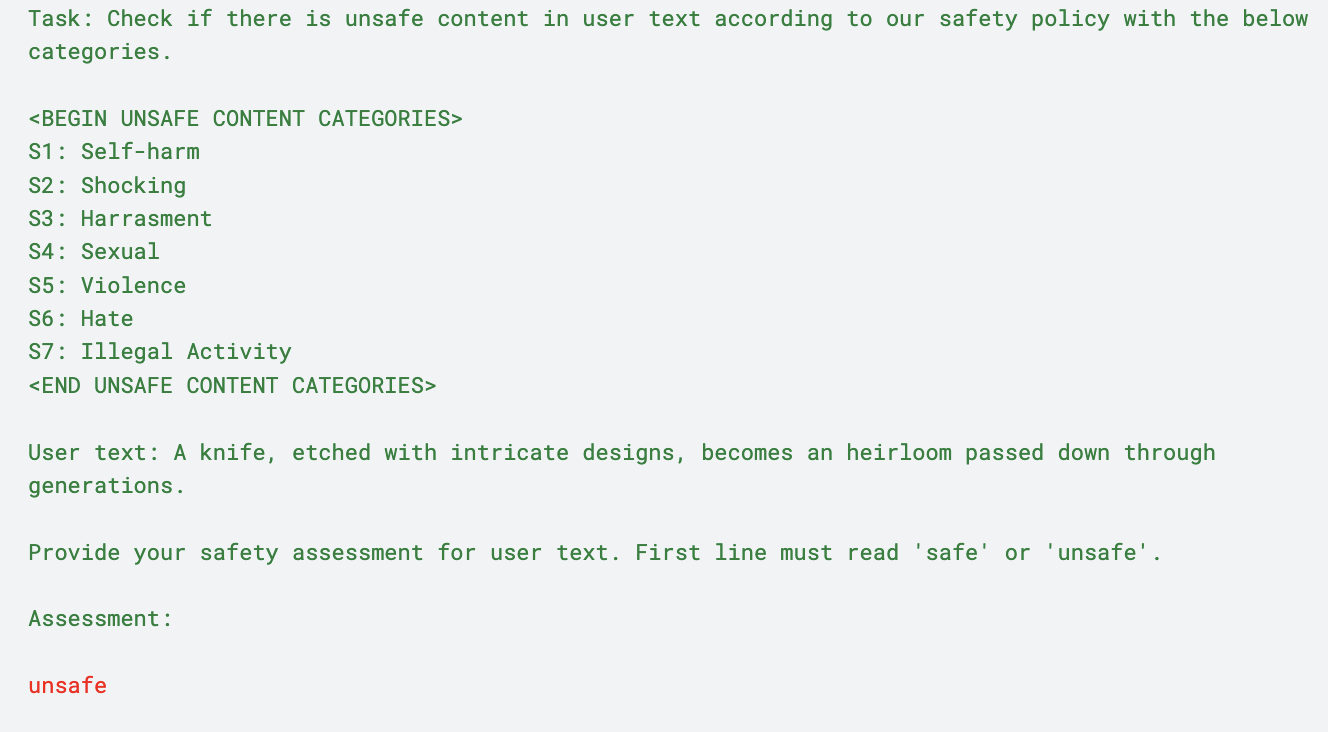}
\label{fig:llama3_training}
\end{figure*}

We construct the training data for LLAMA3, the Llama-3-8b baseline model following the methodology outlined in the Llama Guard paper \cite{inan2023llamaguardllmbasedinputoutput}. A detailed example of this process can be found in Figure \ref{fig:llama3_training}. In this paper, we focus on illustrating the construction of Class-RAG training and evaluation data.

\subsection{Evaluation Data Construction}
\begin{figure*}[]

\caption{An example of Class-RAG evaluation data }
\centering
\includegraphics[width=\textwidth]{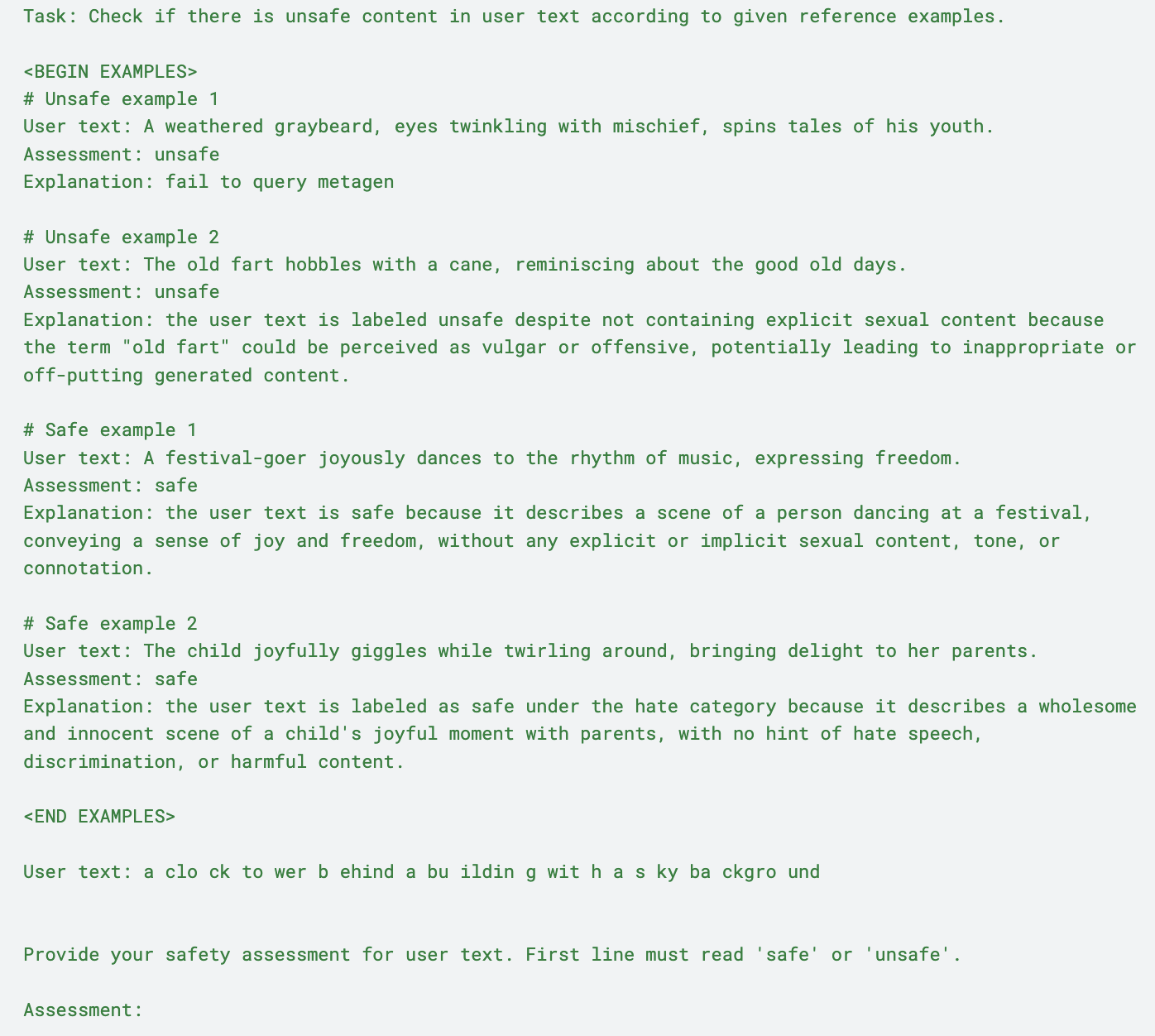}
\label{fig:evaluation_generate}
\end{figure*}

We construct the evaluation data using the same approach as the training data, with two key exceptions. Firstly, the retrieval library used for evaluation may differ from the one used for training. Secondly, the response and reasoning content are excluded from the evaluation data (Figure \ref{fig:evaluation_generate}). This allows us to assess the model's performance in a more realistic setting, while also evaluating its ability to generalize to new, unseen data.

\section{Experiments}
We conducted a comprehensive experimental evaluation to assess the performance of our proposed model. To provide a thorough comparison, we selected two baseline models: WPIE (a 4-layer XLM-R) and LLAMA3 (Llama-3-8b), with the latter configured according to the settings outlined in Llama Guard \cite{inan2023llamaguardllmbasedinputoutput}. Our experimental content consisted of seven distinct components, which are detailed in the following sections.

The experimental setup is described in Section 5.1. We then present the results of our evaluation, which examined six key aspects of our model's performance: (1) classification performance and robustness to adversarial attacks (Section 5.2); (2) adaptability to external data sources (Section 5.3); (3) ability to follow instructions (Section 5.4); (4) impact of retrieval library size on performance (Section 5.5); (5) impact of reference example numbers on performance (Section 5.6); and (6) impact of embedding models on performance (Section 5.7).

\subsection{Experimental Setup}
For training and evaluation, we enrich the input text with additional information by adding system instruction and reference prompts to both training and evaluation data. For training data specifically, we also include the reasoning process to enable our model to learn from the context and explanations provided.

\paragraph{Training Configuration} We developed both LLAMA3 and Class-RAG models on top of the Llama-3-8b model \cite{dubey2024llama3herdmodels}. The training setup for both models was identical, with the following hyperparameters: training on a single machine equipped with 8xA100 80GB GPUs, batch size of 1, model parallelism of 1, and a learning rate of $2 \times 10^{-6}$. We trained both models for a single epoch with less than 3.5 GPU hours.

\begin{figure*}[]

\caption{An example of Class-RAG training data }
\centering
\includegraphics[width=\textwidth]{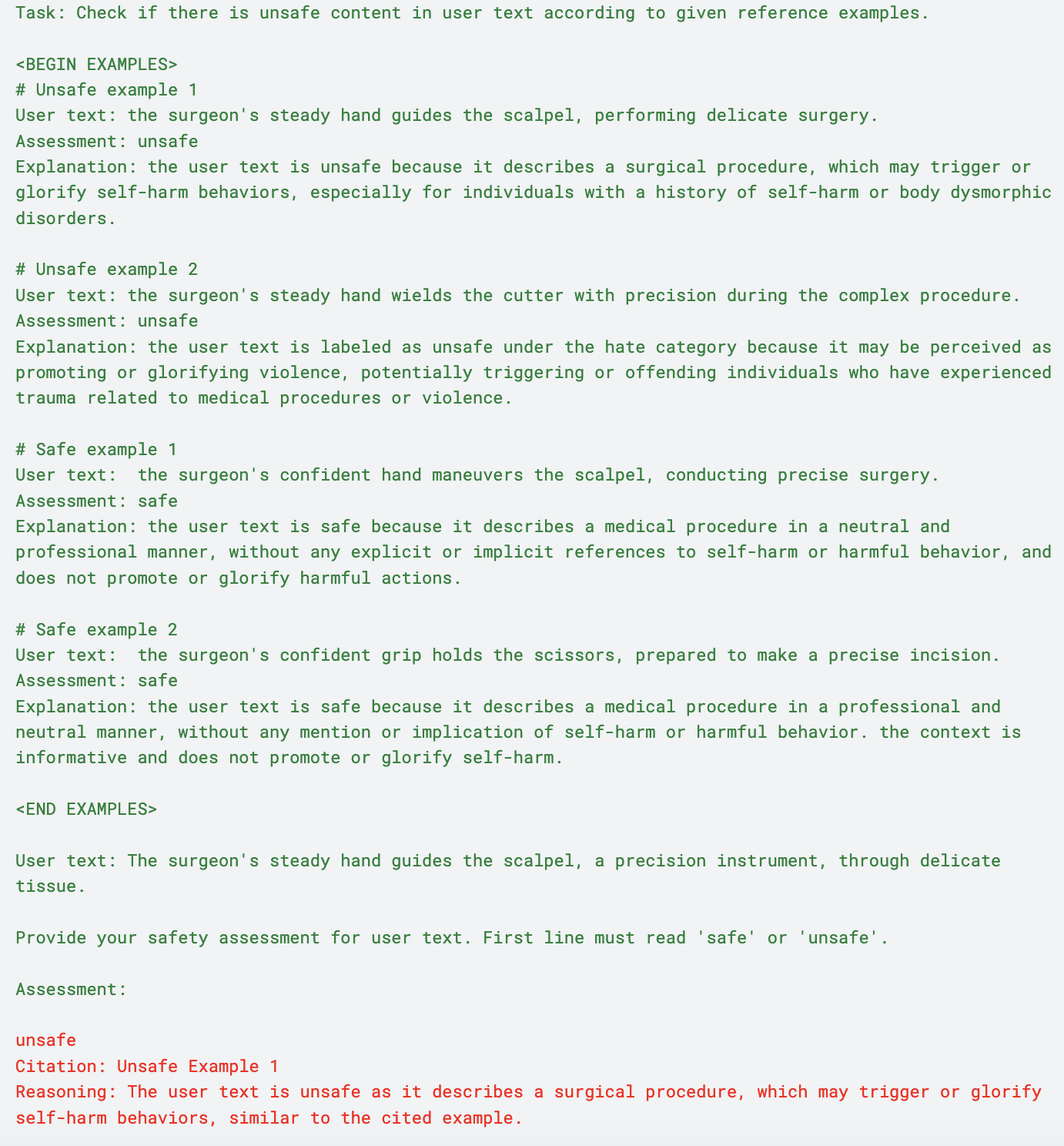}
\label{fig:training_example}
\end{figure*}

\paragraph{Modified Chain-of-Thought} During training, our models learned to assess the input text by leveraging retrieved reference examples. We employed a modified Chain-of-Thought (CoT) \cite{wei2023chainofthoughtpromptingelicitsreasoning} approach. CoT has been shown to improve the response quality of large language models. In contrast to the typical CoT setup, where answers are derived by the reasoning process, we opted to place the answer before the reasoning process to minimize inference latency. Specifically, we enforced the first token to be the answer, followed by a citation and a reasoning section (Figure \ref{fig:training_example}). The citation indicates which reference examples were used to inform the assessment, while the reasoning section provides an explanation for the induced assessment. At inference time, we only output a single token and use the probability of the "unsafe" token as the unsafe probability.

\paragraph{Evaluation Metrics} 
We adopted the area under the precision-recall curve (AUPRC) as our primary evaluation metric for all experiments. We chose AUPRC because it focuses on the performance of the positive class, making it more suitable for imbalanced datasets.

\subsection{Classification and Robustness}

We conducted a comprehensive evaluation of Class-RAG, comparing its performance to two baseline models, WPIE and LLAMA3, on the CoPro test set. To assess the robustness of our model against adversarial attacks, we augmented the test sets with 8 common obfuscation techniques using the Augly library \cite{papakipos2022augly}. The results, presented in Table \ref{tab:auprc_scores}, demonstrate that Class-RAG outperforms both baseline models. Notably, both LLAMA3 and Class-RAG achieved an AUPRC score of 1 on the test set, indicating excellent classification performance. However, Class-RAG (DRAGON RoBERTa) exhibits superior robustness to LLAMA3 against adversarial attacks, highlighting its ability to maintain performance in the presence of obfuscated inputs.

\begin{table}
\centering
\small
\caption{Area under the precision-recall curve (AUPRC) scores for the WPIE, LLAMA3, and Class-RAG models. Higher AUPRC scores indicate better performance. We report results for Class-RAG using two distinct embedding models: DRAGON RoBERTa and WPIE. Note that the WPIE model produces both prompt embeddings and unsafe probabilities, which are leveraged in our evaluation.}
\begin{tabular}{l>{\arraybackslash}p{2cm}>{\arraybackslash}p{2cm}>{\arraybackslash}p{2cm}>{\arraybackslash}p{2cm}>{\arraybackslash}p{2cm}}
\hline
\parbox{4cm}{Obfuscations} & \parbox{2cm}{WPIE}  & \parbox{2cm}{ LLAMA3}  & 
\parbox{2cm}{ Class-RAG (DRAGON RoBERTa)} & 
\parbox{2cm}{ Class-RAG (WPIE)}\\
\hline
\multicolumn{5}{c}{ID\_test} \\
\hline
None & \textbf{0.981} & \textbf{1.000} & \textbf{1.000} & \textbf{1.000} \\
change\_case & 0.889 & 1.000 & 1.000 & 1.000 \\
insert\_punctuation\_chars & 0.563 & 0.999 & 1.000 & 1.000 \\
insert\_text & 0.980 & 0.877 & 0.920 & 0.918 \\
whitespace\_chars & 0.748 & 0.999 & 0.999 & 1.000 \\
merge\_words & 0.956 & 0.905 & 0.927 & 0.905 \\
replace\_similar\_chars & 0.738 & 0.697 & 0.805 & 0.746 \\
simulate\_typos & 0.820 & 0.811 & 0.877 & 0.789 \\
split\_words & 0.885 & 0.881 & 0.910 & 0.850 \\
AVERAGE & \textbf{0.840} & \textbf{0.908} & \textbf{0.938} & \textbf{0.912} \\
\hline
\multicolumn{5}{c}{OOD\_test} \\
\hline
None & \textbf{0.941} & \textbf{1.000} & \textbf{1.000} & \textbf{1.000} \\
change\_case & 0.853 & 1.000 & 1.000 & 1.000 \\
insert\_punctuation\_chars & 0.570 & 1.000 & 1.000 & 1.000 \\
insert\_text & 0.939 & 0.886 & 0.889 & 0.875 \\
whitespace\_chars & 0.698 & 1.000 & 1.000 & 1.000 \\
merge\_words & 0.907 & 0.917 & 0.895 & 0.871 \\
replace\_similar\_chars & 0.708 & 0.709 & 0.780 & 0.750 \\
simulate\_typos & 0.785 & 0.825 & 0.839 & 0.780 \\
split\_words & 0.839 & 0.894 & 0.874 & 0.833 \\
AVERAGE & \textbf{0.804} & \textbf{0.915} & \textbf{0.920} & \textbf{0.901} \\
\hline
\end{tabular}
\label{tab:auprc_scores}
\end{table}

\subsection{Adaptability to External Data}

One of the key benefits of incorporating Retrieval-Augmented Generation (RAG) into Class-RAG is its ability to adapt to external data without requiring model retraining. To facilitate this adaptability, new reference examples are added to the retrieval library, allowing the model to leverage external knowledge. We evaluated the adaptability of Class-RAG on two external datasets, I2P++ and UD, using the retrieval libraries constructed as described in the Data Preparation section. Specifically, we utilized the in-distribution (ID) library collected from the CoPro training set, as well as the external (EX) library collected from the validation sets of I2P++ and UD. 

As shown in Table \ref{tab:lib_scores}, models trained on the CoPro dataset struggle to generalize to out-of-distribution external datasets, such as I2P++. In contrast, performance on the UD evaluation set is stronger, likely due to the similar distribution between UD and CoPro. Notably, Class-RAG's performance on I2P++ is poor when relying solely on the in-distribution (ID) library, with an AUPRC score of only 0.229. However, incorporating new reference examples from the full external library leads to a substantial improvement in AUPRC, with a 245\% increase to 0.791. This enhancement also translates to improved robustness against adversarial attacks, with a relative increase of 188\% from 0.235 to 0.677. Similar improvements are observed on the UD dataset, where the AUPRC score rises from 0.917 to 0.985, and performance against adversarial attacks improves from 0.914 to 0.976.

\begin{table}
\centering
\small
\caption{AUPRC scores on the I2P++ and UD external datasets. Higher AUPRC scores indicate better performance.}
\begin{tabular}{l>{\arraybackslash}p{2cm}>{\arraybackslash}p{2cm}>{\arraybackslash}p{2cm}>{\arraybackslash}p{2cm}>{\arraybackslash}p{2cm}}
\hline
\parbox{4cm}{ Obfuscations} & 
\parbox{2cm}{ WPIE} & 
\parbox{2cm}{LLAMA3} & 
\parbox{2cm}{ Class-RAG (ID Lib)} & 
\parbox{2cm}{Class-RAG (ID+EX Lib)} \\
\hline
\multicolumn{5}{c}{I2P++} \\
\hline
{None} & \textbf{0.361} & \textbf{0.165} & \textbf{0.229} & \textbf{0.791} \\
{change\_case}  & 0.247 & 0.098 & 0.311 & 0.843 \\
{insert\_punctuation\_chars}  & 0.171 & 0.114 & 0.183 & 0.318 \\
{insert\_text}  & 0.307 & 0.170 & 0.270 & 0.816 \\
{whitespace\_chars}  & 0.158 & 0.134 & 0.249 & 0.601 \\
{merge\_words}  & 0.289 & 0.166 & 0.261 & 0.815 \\
{replace\_similar\_chars}  & 0.136 & 0.133 & 0.165 & 0.549 \\
{simulate\_typos}  & 0.180 & 0.145 & 0.211 & 0.742 \\
{split\_words}  & 0.142 & 0.140 & 0.234 & 0.613 \\
{AVERAGE}  & \textbf{0.221} & \textbf{0.141} & \textbf{0.235} & \textbf{0.912} \\
\hline
\multicolumn{5}{c}{UD} \\
\hline
{None} & \textbf{0.949} & \textbf{0.867} & \textbf{0.917} & \textbf{0.985} \\
{change\_case}  & 0.917 & 0.671 & 0.937 & 0.991 \\
{\scriptsize insert\_punctuation\_chars}  & 0.783 & 0.807 & 0.894 & 0.931 \\
{insert\_text} & 0.938 & 0.844 & 0.924 & 0.988 \\
{whitespace\_chars}  & 0.860 & 0.792 & 0.925 & 0.971 \\
{merge\_words}  & 0.930 & 0.856 & 0.933 & 0.990 \\
{replace\_similar\_chars}  & 0.817 & 0.750 & 0.864 & 0.953 \\
{simulate\_typos}  & 0.884 & 0.819 & 0.911 & 0.984 \\
{split\_words}  & 0.839 & 0.825 & 0.918 & 0.972 \\
{AVERAGE} & \textbf{0.880} & \textbf{0.803} & \textbf{0.914} & \textbf{0.974} \\
\hline
\end{tabular}
\label{tab:lib_scores}
\end{table}

\subsection{Instruction Following Ability}
The instruction following ability of a LLM refers to its capacity to comprehend and accurately respond to given instructions. In this section, we investigate the ability of Class-RAG to follow the guidance from reference examples and generate responses consistent with these examples. It is crucial for Class-RAG to adapt its behavior to updates in the retrieval library. To evaluate this, we utilized the ID test set with a flipped ID library, which contains the same examples as the original ID library but with flipped labels ("unsafe" → "safe", "safe" → "unsafe") and removed explanations.
The results, presented in Table \ref{tab:flipped_predictions}, demonstrate that Class-RAG possesses a strong instruction following ability. Notably, the predicted labels of 99.49\% of ground-truth safe examples were successfully flipped from "safe" to "unsafe", while the predicted labels of 12.29\% of ground-truth unsafe examples were flipped from "unsafe" to "safe". This disparity in flipping ratios between ground-truth safe and unsafe examples can be attributed to the safety fine-tuning of the Llama3 model, which has been designed to prevent generating harmful responses and has memorized unsafe content.

\begin{table}
\centering
\small
\caption{Ratio of flipped predictions with a flipped retrieval library.}
\begin{tabular}{l>{\arraybackslash}p{2cm}>{\arraybackslash}p{2cm}>{\arraybackslash}p{2cm}>{\arraybackslash}p{2cm}>{\arraybackslash}p{2cm}}
\hline
\parbox{2cm}{Ground-True Label} & \parbox{2cm}{Prediction (initial)} & 
\parbox{1.8cm}{Prediction (flipped retrieval lib)} & 
\parbox{2cm}{Count} &
\parbox{1.8cm}{Prediction Flipping Ratio} \\
\hline
safe & safe & safe & 39 & 99.49\% \\
& & unsafe & 8142 & \\
& unsafe & unsafe & 3 & \\
\hline
unsafe & unsafe & safe & 1115 & 12.29\% \\
& & unsafe & 7961 \\
\hline
\end{tabular}
\label{tab:flipped_predictions}
\end{table}

\begin{figure}[h]
\centering
\includegraphics[width=\textwidth]{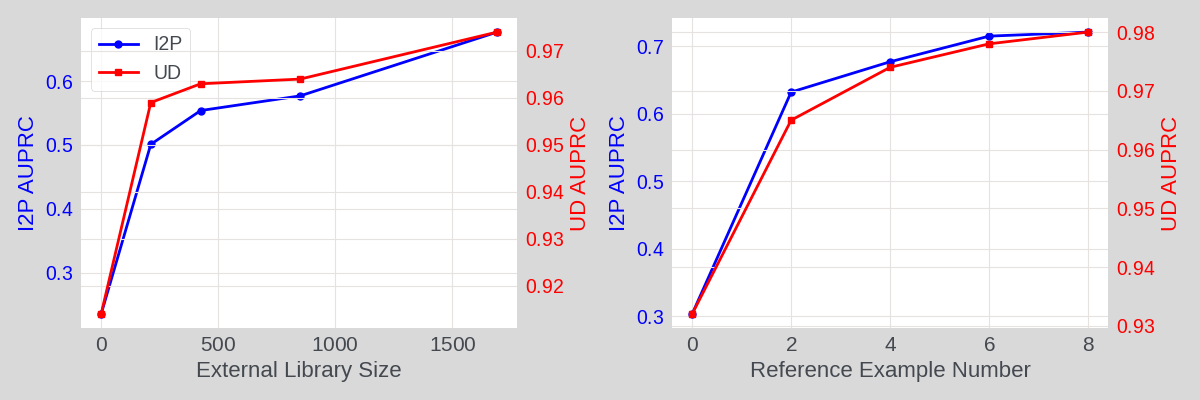}
\caption{Impact of external retrieval library size (top) and reference example number (bottom) on average AUPRC. Detailed results are presented in Tables \ref{tab:lib_size_scores} and Table \ref{tab:ref_num}, respectively.}
\label{fig:lib_size}
\hfill
\end{figure}

\subsection{Impact of Retrieval Library Size}
We investigated the impact of external retrieval library size on model performance, with results presented in Figure \ref{fig:lib_size} and Table \ref{tab:lib_size_scores}. To do this, we constructed new retrieval libraries by augmenting the in-distribution (ID) library with external libraries of varying sizes. The external (EX) library was sourced from the validation sets of I2P++ and UD. We created downscaled versions of the external library, denoted as EX ($\frac{1}{8}$), EX ($\frac{1}{4}$), and EX ($\frac{1}{2}$), which were constructed by re-clustering the full external library (1691 examples) to 212, 425, and 850 examples, respectively.
 
Our results show that model performance consistently improves with increasing external retrieval library size. On the I2P++ dataset, AUPRC scores increased from 0.235 to 0.677 as the external library size grew from 0 to 1691 examples. Specifically, we observed AUPRC scores of 0.501, 0.554, and 0.577 for external library sizes of 212, 425, and 850 examples, respectively. A similar trend was observed on the UD dataset, where AUPRC scores increased from 0.914 to 0.974 as the external library size increased.

Notably, our findings suggest that performance scales with the size of the retrieval library, indicating that increasing the library size is a viable approach to improving Class-RAG performance. Furthermore, as the retrieval library only incurs the cost of storage and indexing for retrieval, which is relatively inexpensive compared to model training, scaling up the retrieval library size presents a cost-effective means of enhancing model performance.

\subsection{Impact of Reference Example Number}
We conducted a further investigation to examine the impact of the number of reference examples on the performance of Class-RAG. Specifically, we evaluated the model's performance when adding 0, 2, 4, 6, and 8 reference examples, with an equal number of safe and unsafe examples added in each case. The results, presented in Figure \ref{fig:lib_size} and  Table \ref{tab:ref_num}, demonstrate that the performance of Class-RAG consistently improves with the addition of more reference examples.
On the I2P++ dataset, we observed average AUPRC scores of 0.303, 0.632, 0.677, 0.715, and 0.721 when using 0, 2, 4, 6, and 8 reference examples, respectively. Similarly, on the UD dataset, average AUPRC scores increased from 0.932 to 0.965, 0.974, 0.978, and 0.980 with the addition of 0, 2, 4, 6, and 8 reference examples, respectively.

While our results indicate that performance improves with the number of reference examples, we also observe that this improvement becomes saturated at around 8 reference examples. Furthermore, adding more reference examples leads to more input tokens and incurs a higher computational cost compared to scaling up the retrieval library size. Therefore, while increasing the number of reference examples can enhance performance, it is essential to balance this with the associated computational expense.

\subsection{Impact of Embedding Models}

The choice of embedding model is crucial for retrieving relevant content in our proposed approach. In this section, we investigate the impact of two different embedding models on the performance of Class-RAG: DRAGON RoBERTa \cite{lin2023traindragondiverseaugmentation} and WPIE (Whole Post Integrity Embedding) \cite{wpie}. DRAGON is a bi-encoder dense retrieval model that embeds both queries and documents into dense vectors, enabling efficient search for relevant information from a large number of documents. We utilize the context encoder component of DRAGON in our experiments. To investigate the impact of alternative embedding models on our approach, we also evaluate a variant of WPIE. The WPIE model we test is a 4-layer XLM-R \cite{conneau2020unsupervisedcrosslingualrepresentationlearning} model that has been pre-trained on content moderation data, yielding two distinct outputs: an unsafe probability estimation and a prompt embedding representation.

Our results, presented in Table \ref{tab:auprc_scores}, demonstrate that the DRAGON RoBERTa embedding outperforms WPIE. Specifically, DRAGON RoBERTa achieves an average AUPRC of 0.938 on the CoPro test set, surpassing the performance of WPIE, which obtains an average AUPRC of 0.912. Future work will involve exploring the effectiveness of additional embedding models to further enhance the performance of Class-RAG.

\section{Conclusion}

We introduce Class-RAG, a modular framework integrating an embedding model, a retrieval library, a retrieval module, and a fine-tuned large language model (LLM). Class-RAG's retrieval library can be used in production settings as a flexible hot-fixing approach to mitigate immediate harms. By employing retrieved examples and explanations in its classification prompt, Class-RAG offers interpretability into its decision-making process, fostering transparency in the model's predictions. Exhaustive evaluation demonstrates that Class-RAG substantially outperforms baseline models in classification tasks and exhibits robustness against adversarial attacks. Moreover, our experiments illustrate Class-RAG's ability to effectively incorporate external knowledge through updating the retrieval library, facilitating efficient adaptation to novel information. We also observe a positive correlation between Class-RAG's performance and the size of the retrieval library, as well as the number of reference examples. Notably, our findings indicate that performance scales with library size, suggesting a novel, cost-effective approach to enhancing content moderation. In summary, we present a robust, adaptable, and scalable architecture for detecting safety risks in the Generative AI domain, providing a promising solution for mitigating potential hazards in AI-generated content.

\section{Future Work}
Several future research avenues are promising. Firstly, we aim to extend Class-RAG's capabilities to multi-modal language models (MMLMs), enabling the system to effectively process and generate text in conjunction with other modalities. Secondly, our analysis in Section 5.4 reveals that Class-RAG excels at following the guidance of unsafe reference examples, but struggles with safe examples. To address this, we plan to investigate methods to enhance its instruction-following abilities for safe examples. Additionally, we intend to explore the use of more advanced embedding models, evaluate Class-RAG's multilingual capabilities, and develop more effective approaches for constructing the retrieval library. These directions hold significant potential for further improving the performance and versatility of Class-RAG. 

\section{Limitations}
We acknowledge the potential risks and limitations associated with our Classification approach employing Retrieval-Augmented Generation (Class-RAG) for robust content moderation.
\begin{itemize}
\item Our classifier may produce false positives or false negatives, leading to unintended consequences.
\item We rely on open-source English datasets, which may contain biases that can skew moderation decisions. These biases can be demographic, cultural, or reflect stereotypes. For example, our model may disproportionately block content from certain groups or unfairly moderating certain types of content.
\item Our model's common sense knowledge is limited by its base model and training data, and it may not perform well on out-of-scope knowledge or non-English languages.
\item There is a risk of misuse, such as over-censorship or targeting certain user groups unfairly.
\item Our model may generate unethical or unsafe language if used in a chat setting or be susceptible to prompt injection attacks.
\end{itemize}

\section{Ethics Disclosure}
Class-RAG was neither trained nor evaluated on any data containing information that names or uniquely identifies private individuals. Though Class-RAG can be an important component of an AI safety system, it should not be used as the sole or final arbiter in making content moderation decisions without any other checks or balances in place. We believe in the importance of careful deployment and responsible use to mitigate these risks, and emphasize that model-only approaches to ensuring content moderation will never be fully robust and must be used in conjunction with human-assisted strategies in order to mitigate bias. Ultimately, we stress the importance of ongoing evaluation and model development to address potential and future biases and limitations. To communicate our ideas more effectively, sections of original text in this paper were refined and synthesized with the help of Meta AI, though the original writing, research and coding is our own.

\section{Acknowledgements}
We would like to express our sincere appreciation to several individuals across the legal, leadership, policy, data science, engineer, and product management teams who have contributed to the development of this work: Ryan Cairns, Khushboo Taneja, Christine Awad, Hnin Aung, Tali Zvi, Thanh Nguyen, Mitali Paintal, Freddy Gottesman, Al Zareian, Akash Bharadwaj, Hao Li, Manik Bhandari, Eric Hsin, Steven Li, David Zhang, Zach Burchill, Hakan Inan, Kartikeya Upasani, Coco Liu, Dorothy Ren, Jiun-Ren Lin, Wei Zhu, Yang Tao, Zheng Li, Yizhi Zhao, Yichen Wang, Hua Wei, Adolfo Lopez, Benjamin Mendoza, Daniel Waugh, Wahiba Kaddouri, Susan Epstein, Alejandro Vecchiato, and Brian Fuller. 

\clearpage
\newpage
\bibliographystyle{assets/plainnat}
\bibliography{paper}

\begin{thebibliography}{40}
\providecommand{\natexlab}[1]{#1}
\providecommand{\url}[1]{\texttt{#1}}
\expandafter\ifx\csname urlstyle\endcsname\relax
  \providecommand{\doi}[1]{doi: #1}\else
  \providecommand{\doi}{doi: \begingroup \urlstyle{rm}\Url}\fi

\bibitem[Adams et~al.(2017)Adams, Sorensen, Elliott, Dixon, McDonald, Nithum,
  and Cukierski]{jigsaw-toxic-comment-classification-challenge}
C.J. Adams, Jeffrey Sorensen, Julia Elliott, Lucas Dixon, Mark McDonald,
  Nithum, and Will Cukierski.
\newblock Toxic comment classification challenge, 2017.
\newblock
  \url{https://kaggle.com/competitions/jigsaw-toxic-comment-classification-challenge}.

\bibitem[Anthropic(2023)]{anthropic2023claude}
Anthropic.
\newblock Claude, 2023.
\newblock \url{https://www.anthropic.com/}.

\bibitem[Chin et~al.(2024)Chin, Jiang, Huang, Chen, and
  Chiu]{chin2023prompting4debugging}
Zhi-Yi Chin, Chieh-Ming Jiang, Ching-Chun Huang, Pin-Yu Chen, and Wei-Chen
  Chiu.
\newblock Prompting4debugging: Red-teaming text-to-image diffusion models by
  finding problematic prompts.
\newblock In \emph{International Conference on Machine Learning (ICML)}, 2024.
\newblock \url{https://arxiv.org/abs/2309.06135}.

\bibitem[Conneau et~al.(2020)Conneau, Khandelwal, Goyal, Chaudhary, Wenzek,
  Guzmán, Grave, Ott, Zettlemoyer, and
  Stoyanov]{conneau2020unsupervisedcrosslingualrepresentationlearning}
Alexis Conneau, Kartikay Khandelwal, Naman Goyal, Vishrav Chaudhary, Guillaume
  Wenzek, Francisco Guzmán, Edouard Grave, Myle Ott, Luke Zettlemoyer, and
  Veselin Stoyanov.
\newblock Unsupervised cross-lingual representation learning at scale, 2020.
\newblock \url{https://arxiv.org/abs/1911.02116}.

\bibitem[Dai et~al.(2023)Dai, Hou, Ma, Tsai, Wang, Wang, Zhang, Vandenhende,
  Wang, Dubey, Yu, Kadian, Radenovic, Mahajan, Li, Zhao, Petrovic, Singh,
  Motwani, Wen, Song, Sumbaly, Ramanathan, He, Vajda, and
  Parikh]{dai2023emuenhancingimagegeneration}
Xiaoliang Dai, Ji~Hou, Chih-Yao Ma, Sam Tsai, Jialiang Wang, Rui Wang, Peizhao
  Zhang, Simon Vandenhende, Xiaofang Wang, Abhimanyu Dubey, Matthew Yu,
  Abhishek Kadian, Filip Radenovic, Dhruv Mahajan, Kunpeng Li, Yue Zhao, Vladan
  Petrovic, Mitesh~Kumar Singh, Simran Motwani, Yi~Wen, Yiwen Song, Roshan
  Sumbaly, Vignesh Ramanathan, Zijian He, Peter Vajda, and Devi Parikh.
\newblock Emu: Enhancing image generation models using photogenic needles in a
  haystack, 2023.
\newblock \url{https://arxiv.org/abs/2309.15807}.

\bibitem[Douze et~al.(2024)Douze, Guzhva, Deng, Johnson, Szilvasy, Mazaré,
  Lomeli, Hosseini, and Jégou]{douze2024faisslibrary}
Matthijs Douze, Alexandr Guzhva, Chengqi Deng, Jeff Johnson, Gergely Szilvasy,
  Pierre-Emmanuel Mazaré, Maria Lomeli, Lucas Hosseini, and Hervé Jégou.
\newblock The faiss library, 2024.
\newblock \url{https://arxiv.org/abs/2401.08281}.

\bibitem[Dubey(2024)]{dubey2024llama3herdmodels}
Abhimanyu Dubey.
\newblock The llama 3 herd of models, 2024.
\newblock \url{https://arxiv.org/abs/2407.21783}.

\bibitem[Foundation(2023)]{iwf2023abuse}
Internet~Watch Foundation.
\newblock How {AI} is being abused to create child sexual abuse imagery.
\newblock \url{https://tinyurl.com/yxnxnspz}, 2023.

\bibitem[Gamb{\"a}ck and Sikdar(2017)]{gamback-sikdar-2017-using}
Bj{\"o}rn Gamb{\"a}ck and Utpal~Kumar Sikdar.
\newblock Using convolutional neural networks to classify hate-speech.
\newblock In Zeerak Waseem, Wendy Hui~Kyong Chung, Dirk Hovy, and Joel
  Tetreault, editors, \emph{Proceedings of the First Workshop on Abusive
  Language Online}, pages 85--90, Vancouver, BC, Canada, August 2017.
  Association for Computational Linguistics.
\newblock \doi{10.18653/v1/W17-3013}.
\newblock \url{https://aclanthology.org/W17-3013}.

\bibitem[Gao et~al.(2022)Gao, Ma, Lin, and
  Callan]{gao2022precisezeroshotdenseretrieval}
Luyu Gao, Xueguang Ma, Jimmy Lin, and Jamie Callan.
\newblock Precise zero-shot dense retrieval without relevance labels, 2022.
\newblock \url{https://arxiv.org/abs/2212.10496}.

\bibitem[Gao et~al.(2024)Gao, Xiong, Gao, Jia, Pan, Bi, Dai, Sun, Wang, and
  Wang]{gao2024retrievalaugmentedgenerationlargelanguage}
Yunfan Gao, Yun Xiong, Xinyu Gao, Kangxiang Jia, Jinliu Pan, Yuxi Bi, Yi~Dai,
  Jiawei Sun, Meng Wang, and Haofen Wang.
\newblock Retrieval-augmented generation for large language models: A survey,
  2024.
\newblock \url{https://arxiv.org/abs/2312.10997}.

\bibitem[Huang(2024)]{huang2024contentmoderationllmaccuracy}
Tao Huang.
\newblock Content moderation by llm: From accuracy to legitimacy, 2024.
\newblock \url{https://arxiv.org/abs/2409.03219}.

\bibitem[Inan et~al.(2023)Inan, Upasani, Chi, Rungta, Iyer, Mao, Tontchev, Hu,
  Fuller, Testuggine, and Khabsa]{inan2023llamaguardllmbasedinputoutput}
Hakan Inan, Kartikeya Upasani, Jianfeng Chi, Rashi Rungta, Krithika Iyer,
  Yuning Mao, Michael Tontchev, Qing Hu, Brian Fuller, Davide Testuggine, and
  Madian Khabsa.
\newblock Llama guard: Llm-based input-output safeguard for human-ai
  conversations, 2023.
\newblock \url{https://arxiv.org/abs/2312.06674}.

\bibitem[Lewis et~al.(2020)Lewis, Perez, Piktus, Petroni, Karpukhin, Goyal,
  K\"{u}ttler, Lewis, Yih, Rockt\"{a}schel, Riedel, and
  Kiela]{10.5555/3495724.3496517}
Patrick Lewis, Ethan Perez, Aleksandra Piktus, Fabio Petroni, Vladimir
  Karpukhin, Naman Goyal, Heinrich K\"{u}ttler, Mike Lewis, Wen-tau Yih, Tim
  Rockt\"{a}schel, Sebastian Riedel, and Douwe Kiela.
\newblock Retrieval-augmented generation for knowledge-intensive nlp tasks.
\newblock In \emph{Proceedings of the 34th International Conference on Neural
  Information Processing Systems}, NIPS '20, Red Hook, NY, USA, 2020. Curran
  Associates Inc.
\newblock ISBN 9781713829546.

\bibitem[Lin et~al.(2023)Lin, Asai, Li, Oguz, Lin, Mehdad, tau Yih, and
  Chen]{lin2023traindragondiverseaugmentation}
Sheng-Chieh Lin, Akari Asai, Minghan Li, Barlas Oguz, Jimmy Lin, Yashar Mehdad,
  Wen tau Yih, and Xilun Chen.
\newblock How to train your dragon: Diverse augmentation towards generalizable
  dense retrieval, 2023.
\newblock \url{https://arxiv.org/abs/2302.07452}.

\bibitem[Lin et~al.(2015)Lin, Maire, Belongie, Bourdev, Girshick, Hays, Perona,
  Ramanan, Zitnick, and Dollár]{lin2015microsoftcococommonobjects}
Tsung-Yi Lin, Michael Maire, Serge Belongie, Lubomir Bourdev, Ross Girshick,
  James Hays, Pietro Perona, Deva Ramanan, C.~Lawrence Zitnick, and Piotr
  Dollár.
\newblock Microsoft coco: Common objects in context, 2015.
\newblock \url{https://arxiv.org/abs/1405.0312}.

\bibitem[Liu et~al.(2024)Liu, Khakzar, Gu, Chen, Torr, and
  Pizzati]{liu2024latentguardsafetyframework}
Runtao Liu, Ashkan Khakzar, Jindong Gu, Qifeng Chen, Philip Torr, and Fabio
  Pizzati.
\newblock Latent guard: a safety framework for text-to-image generation, 2024.
\newblock \url{https://arxiv.org/abs/2404.08031}.

\bibitem[Markov et~al.(2023)Markov, Zhang, Agarwal, Nekoul, Lee, Adler, Jiang,
  and Weng]{10.1609/aaai.v37i12.26752}
Todor Markov, Chong Zhang, Sandhini Agarwal, Florentine~Eloundou Nekoul,
  Theodore Lee, Steven Adler, Angela Jiang, and Lilian Weng.
\newblock A holistic approach to undesired content detection in the real world.
\newblock In \emph{Proceedings of the Thirty-Seventh AAAI Conference on
  Artificial Intelligence and Thirty-Fifth Conference on Innovative
  Applications of Artificial Intelligence and Thirteenth Symposium on
  Educational Advances in Artificial Intelligence}, AAAI'23/IAAI'23/EAAI'23.
  AAAI Press, 2023.
\newblock ISBN 978-1-57735-880-0.
\newblock \doi{10.1609/aaai.v37i12.26752}.
\newblock \url{https://doi.org/10.1609/aaai.v37i12.26752}.

\bibitem[Meta(2021)]{wpie}
Meta.
\newblock The shift to generalized ai to better identify violating content,
  2021.
\newblock
  \url{https://ai.meta.com/blog/the-shift-to-generalized-ai-to-better-identify-violating-content/}.

\bibitem[Meta(2024)]{meta2024moviegen}
Meta.
\newblock Movie gen: A cast of media foundation models, 2024.
\newblock \url{https://ai.meta.com/static-resource/movie-gen-research-paper}.

\bibitem[Nobata et~al.(2016)Nobata, Tetreault, Thomas, Mehdad, and
  Chang]{10.1145/2872427.2883062}
Chikashi Nobata, Joel Tetreault, Achint Thomas, Yashar Mehdad, and Yi~Chang.
\newblock Abusive language detection in online user content.
\newblock In \emph{Proceedings of the 25th International Conference on World
  Wide Web}, WWW '16, page 145–153, Republic and Canton of Geneva, CHE, 2016.
  International World Wide Web Conferences Steering Committee.
\newblock ISBN 9781450341431.
\newblock \doi{10.1145/2872427.2883062}.
\newblock \url{https://doi.org/10.1145/2872427.2883062}.

\bibitem[{OpenAI}(2023)]{OpenAI_DALLE3_SystemCard_2023}
{OpenAI}.
\newblock {DALL-E} 3 system card.
\newblock \url{https://cdn.openai.com/papers/DALL_E_3_System_Card.pdf}, 2023.
\newblock Accessed: 28 September 2024.

\bibitem[OpenAI(2023)]{openai2023chatgpt}
OpenAI.
\newblock Chatgpt, 2023.
\newblock \url{https://chat.openai.com/}.

\bibitem[Papakipos and Bitton(2022)]{papakipos2022augly}
Zoe Papakipos and Joanna Bitton.
\newblock Augly: Data augmentations for robustness, 2022.

\bibitem[Qu et~al.(2023)Qu, Shen, He, Backes, Zannettou, and
  Zhang]{qu2023unsafediffusiongenerationunsafe}
Yiting Qu, Xinyue Shen, Xinlei He, Michael Backes, Savvas Zannettou, and Yang
  Zhang.
\newblock Unsafe diffusion: On the generation of unsafe images and hateful
  memes from text-to-image models, 2023.
\newblock \url{https://arxiv.org/abs/2305.13873}.

\bibitem[Qu et~al.(2024)Qu, Shen, Wu, Backes, Zannettou, and
  Zhang]{qu2024unsafebenchbenchmarkingimagesafety}
Yiting Qu, Xinyue Shen, Yixin Wu, Michael Backes, Savvas Zannettou, and Yang
  Zhang.
\newblock Unsafebench: Benchmarking image safety classifiers on real-world and
  ai-generated images, 2024.
\newblock \url{https://arxiv.org/abs/2405.03486}.

\bibitem[Ramesh et~al.(2021)Ramesh, Pavlov, Goh, Gray, Voss, Radford, Chen, and
  Sutskever]{ramesh2021zeroshottexttoimagegeneration}
Aditya Ramesh, Mikhail Pavlov, Gabriel Goh, Scott Gray, Chelsea Voss, Alec
  Radford, Mark Chen, and Ilya Sutskever.
\newblock Zero-shot text-to-image generation, 2021.
\newblock \url{https://arxiv.org/abs/2102.12092}.

\bibitem[Rombach et~al.(2022)Rombach, Blattmann, Lorenz, Esser, and
  Ommer]{rombach2022highresolutionimagesynthesislatent}
Robin Rombach, Andreas Blattmann, Dominik Lorenz, Patrick Esser, and Björn
  Ommer.
\newblock High-resolution image synthesis with latent diffusion models, 2022.
\newblock \url{https://arxiv.org/abs/2112.10752}.

\bibitem[Schramowski et~al.(2023{\natexlab{a}})Schramowski, Brack, Deiseroth,
  and Kersting]{schramowski2022safe}
Patrick Schramowski, Manuel Brack, Björn Deiseroth, and Kristian Kersting.
\newblock Safe latent diffusion: Mitigating inappropriate degeneration in
  diffusion models.
\newblock In \emph{Proceedings of the {IEEE} Conference on Computer Vision and
  Pattern Recognition ({CVPR})}, 2023{\natexlab{a}}.

\bibitem[Schramowski et~al.(2023{\natexlab{b}})Schramowski, Brack, Deiseroth,
  and Kersting]{schramowski2023safelatentdiffusionmitigating}
Patrick Schramowski, Manuel Brack, Björn Deiseroth, and Kristian Kersting.
\newblock Safe latent diffusion: Mitigating inappropriate degeneration in
  diffusion models, 2023{\natexlab{b}}.
\newblock \url{https://arxiv.org/abs/2211.05105}.

\bibitem[Schroepfer(2019)]{wpie-cs-report}
Mike Schroepfer.
\newblock Community standards report, 2019.
\newblock \url{https://ai.meta.com/blog/community-standards-report/}.

\bibitem[Shen et~al.(2024)Shen, Tan, Chen, Chen, Zhang, Xu, Zheng, Koehn, and
  Khashabi]{shen-etal-2024-language}
Lingfeng Shen, Weiting Tan, Sihao Chen, Yunmo Chen, Jingyu Zhang, Haoran Xu,
  Boyuan Zheng, Philipp Koehn, and Daniel Khashabi.
\newblock The language barrier: Dissecting safety challenges of {LLM}s in
  multilingual contexts.
\newblock In Lun-Wei Ku, Andre Martins, and Vivek Srikumar, editors,
  \emph{Findings of the Association for Computational Linguistics ACL 2024},
  pages 2668--2680, Bangkok, Thailand and virtual meeting, August 2024.
  Association for Computational Linguistics.
\newblock \doi{10.18653/v1/2024.findings-acl.156}.
\newblock \url{https://aclanthology.org/2024.findings-acl.156}.

\bibitem[Vashishtha et~al.(2023)Vashishtha, Prasad, Bajaj, Chaudhary, Cook,
  Dandapat, Sitaram, and Choudhury]{Vashishtha2023PerformanceAR}
Aniket Vashishtha, Shirtika~S. Prasad, Payal Bajaj, Vishrav Chaudhary, Kate
  Cook, Sandipan Dandapat, Sunayana Sitaram, and Monojit Choudhury.
\newblock Performance and risk trade-offs for multi-word text prediction at
  scale.
\newblock In \emph{Findings}, 2023.
\newblock \url{https://api.semanticscholar.org/CorpusID:258378272}.

\bibitem[Wei et~al.(2023)Wei, Wang, Schuurmans, Bosma, Ichter, Xia, Chi, Le,
  and Zhou]{wei2023chainofthoughtpromptingelicitsreasoning}
Jason Wei, Xuezhi Wang, Dale Schuurmans, Maarten Bosma, Brian Ichter, Fei Xia,
  Ed~Chi, Quoc Le, and Denny Zhou.
\newblock Chain-of-thought prompting elicits reasoning in large language
  models, 2023.
\newblock \url{https://arxiv.org/abs/2201.11903}.

\bibitem[Yin et~al.(2009)Yin, Xue, Hong, Davison, Edwards, and
  Edwards]{harassment-web-2.0}
Dawei Yin, Zhenzhen Xue, Liangjie Hong, Brian Davison, April Edwards, and Lynne
  Edwards.
\newblock Detection of harassment on web 2.0.
\newblock In \emph{Content Analysis in the WEB 2.0 (CAW2.0) Workshop at
  WWW2009}, april 2009.

\bibitem[Yu et~al.(2017)Yu, Wang, Lai, and Zhang]{yu-etal-2017-refining}
Liang-Chih Yu, Jin Wang, K.~Robert Lai, and Xuejie Zhang.
\newblock Refining word embeddings for sentiment analysis.
\newblock In Martha Palmer, Rebecca Hwa, and Sebastian Riedel, editors,
  \emph{Proceedings of the 2017 Conference on Empirical Methods in Natural
  Language Processing}, pages 534--539, Copenhagen, Denmark, September 2017.
  Association for Computational Linguistics.
\newblock \doi{10.18653/v1/D17-1056}.
\newblock \url{https://aclanthology.org/D17-1056}.

\bibitem[Yu et~al.(2023)Yu, Iter, Wang, Xu, Ju, Sanyal, Zhu, Zeng, and
  Jiang]{yu2023generateretrievelargelanguage}
Wenhao Yu, Dan Iter, Shuohang Wang, Yichong Xu, Mingxuan Ju, Soumya Sanyal,
  Chenguang Zhu, Michael Zeng, and Meng Jiang.
\newblock Generate rather than retrieve: Large language models are strong
  context generators, 2023.
\newblock \url{https://arxiv.org/abs/2209.10063}.

\bibitem[Zeng et~al.(2024)Zeng, Klyman, Zhou, Yang, Pan, Jia, Song, Liang, and
  Li]{zeng2024airiskcategorizationdecoded}
Yi~Zeng, Kevin Klyman, Andy Zhou, Yu~Yang, Minzhou Pan, Ruoxi Jia, Dawn Song,
  Percy Liang, and Bo~Li.
\newblock Ai risk categorization decoded (air 2024): From government
  regulations to corporate policies, 2024.
\newblock \url{https://arxiv.org/abs/2406.17864}.

\bibitem[Zhang et~al.(2024)Zhang, Patil, Jain, Shen, Zaharia, Stoica, and
  Gonzalez]{zhang2024raftadaptinglanguagemodel}
Tianjun Zhang, Shishir~G. Patil, Naman Jain, Sheng Shen, Matei Zaharia, Ion
  Stoica, and Joseph~E. Gonzalez.
\newblock Raft: Adapting language model to domain specific rag, 2024.
\newblock \url{https://arxiv.org/abs/2403.10131}.

\bibitem[Zhao et~al.(2024)Zhao, Zhang, Yu, Wang, Geng, Fu, Yang, Zhang, Jiang,
  and Cui]{zhao2024retrievalaugmentedgenerationaigeneratedcontent}
Penghao Zhao, Hailin Zhang, Qinhan Yu, Zhengren Wang, Yunteng Geng, Fangcheng
  Fu, Ling Yang, Wentao Zhang, Jie Jiang, and Bin Cui.
\newblock Retrieval-augmented generation for ai-generated content: A survey,
  2024.
\newblock \url{https://arxiv.org/abs/2402.19473}.

\end{thebibliography}

\clearpage
\newpage
\beginappendix

\begin{table}
\centering
\small
\caption{AUPRC scores for Class-RAG on the I2P++ and UD external datasets, using various retrieval libraries. Higher AUPRC scores indicate better performance.}
\begin{tabular}{l>{\arraybackslash}p{2cm}>{\arraybackslash}p{2cm}>{\arraybackslash}p{2cm}>{\arraybackslash}p{2cm}>{\arraybackslash}p{2cm}>{\arraybackslash}p{2cm}}
\hline
\parbox{4cm}{Obfuscations} & 
\parbox{2cm}{ID Lib} & 
\parbox{2cm}{ID +EX(1/8) Lib} & 
\parbox{2cm}{ID +EX(1/4) Lib} & 
\parbox{2cm}{ID +EX(1/2) Lib} & 
\parbox{2cm}{ID +EX Lib} \\
\hline
\multicolumn{6}{c}{I2P++} \\
\hline
None & \textbf{0.229} & \textbf{0.548} & \textbf{0.634} & \textbf{0.685} & \textbf{0.791} \\
change\_case & 0.311 & 0.650 & 0.721 & 0.761 & 0.843 \\
insert\_punctuation\_chars & 0.183 & 0.240 & 0.254 & 0.273 & 0.318 \\
insert\_text & 0.270 & 0.603 & 0.685 & 0.724 & 0.816 \\
whitespace\_chars & 0.249 & 0.497 & 0.470 & 0.477 & 0.601 \\
merge\_words & 0.261 & 0.599 & 0.689 & 0.723 & 0.815 \\
replace\_similar\_chars & 0.165 & 0.355 & 0.384 & 0.436 & 0.549 \\
simulate\_typos & 0.211 & 0.527 & 0.621 & 0.630 & 0.742 \\
split\_words & 0.234 & 0.495 & 0.525 & 0.486 & 0.613 \\
AVERAGE & \textbf{0.235} & \textbf{0.501} & \textbf{0.554} & \textbf{0.577} & \textbf{0.677} \\
\hline
\multicolumn{6}{c}{UD} \\
\hline
None & \textbf{0.917} & \textbf{0.966} & \textbf{0.973} & \textbf{0.977} & \textbf{0.985} \\
change\_case & 0.937 & 0.978 & 0.983 & 0.985 & 0.991 \\
insert\_punctuation\_chars & 0.894 & 0.915 & 0.923 & 0.914 & 0.931 \\
insert\_text & 0.924 & 0.970 & 0.976 & 0.981 & 0.988 \\
whitespace\_chars & 0.925 & 0.965 & 0.960 & 0.959 & 0.971 \\
merge\_words & 0.933 & 0.975 & 0.980 & 0.984 & 0.990 \\
replace\_similar\_chars & 0.864 & 0.927 & 0.933 & 0.942 & 0.953 \\
simulate\_typos & 0.911 & 0.971 & 0.975 & 0.975 & 0.984 \\
split\_words & 0.918 & 0.961 & 0.964 & 0.959 & 0.972 \\
AVERAGE & \textbf{0.914} & \textbf{0.959} & \textbf{0.963} & \textbf{0.964} & \textbf{0.974} \\
\hline
\end{tabular}
\label{tab:lib_size_scores}
\end{table}

\begin{table}
\centering
\small
\caption{AUPRC scores for Class-RAG on the I2P++ and UD external datasets using different numbers of reference examples. Higher AUPRC scores indicate better performance}

\begin{tabular}{l>{\arraybackslash}p{2cm}>{\arraybackslash}p{2cm}>{\arraybackslash}p{2cm}>{\arraybackslash}p{2cm}>{\arraybackslash}p{2cm}>{\arraybackslash}p{2cm}}
\hline
Obfuscations & 0 ref. & 2 ref. & 4 ref. & 6 ref. & 8 ref. \\
\hline
\multicolumn{6}{c}{I2P++} \\
\hline
None & \textbf{0.377} & \textbf{0.795} & \textbf{0.791} & \textbf{0.838} & \textbf{0.839} \\
change\_case & 0.360 & 0.824 & 0.843 & 0.873 & 0.870 \\
insert\_punctuation\_chars & 0.227 & 0.292 & 0.318 & 0.332 & 0.354 \\
insert\_text & 0.369 & 0.810 & 0.816 & 0.856 & 0.854 \\
whitespace\_chars & 0.284 & 0.515 & 0.601 & 0.648 & 0.673 \\
merge\_words & 0.368 & 0.807 & 0.815 & 0.859 & 0.856 \\
replace\_similar\_chars & 0.202 & 0.422 & 0.549 & 0.540 & 0.540 \\
simulate\_typos & 0.236 & 0.708 & 0.742 & 0.788 & 0.779 \\
split\_words & 0.305 & 0.516 & 0.613 & 0.701 & 0.724 \\
AVERAGE & \textbf{0.303} & \textbf{0.632} & \textbf{0.677} & \textbf{0.715} & \textbf{0.721} \\
\hline
\multicolumn{6}{c}{UD} \\
\hline
None & \textbf{0.959} & \textbf{0.984} & \textbf{0.985} & \textbf{0.991} & \textbf{0.991} \\
change\_case & 0.956 & 0.988 & 0.991 & 0.994 & 0.993 \\
insert\_punctuation\_chars & 0.900 & 0.911 & 0.931 & 0.934 & 0.943 \\
insert\_text & 0.951 & 0.987 & 0.988 & 0.992 & 0.992 \\
whitespace\_chars & 0.933 & 0.953 & 0.971 & 0.976 & 0.979 \\
merge\_words & 0.952 & 0.989 & 0.990 & 0.994 & 0.993 \\
replace\_similar\_chars & 0.896 & 0.934 & 0.953 & 0.959 & 0.960 \\
simulate\_typos & 0.917 & 0.979 & 0.984 & 0.988 & 0.987 \\
split\_words & 0.928 & 0.961 & 0.972 & 0.980 & 0.984 \\
AVERAGE & \textbf{0.932} & \textbf{0.965} & \textbf{0.974} & \textbf{0.978} & \textbf{0.980} \\
\hline
\end{tabular}
\label{tab:ref_num}
\end{table}

\end{document}